\documentclass{llncs}
\usepackage{graphicx}
\usepackage{color}
\usepackage{amsmath}
\usepackage{amssymb}
\usepackage{xspace}
\usepackage{url}

\usepackage[colorlinks,urlcolor=blue]{hyperref}
\usepackage{breakurl}

\usepackage[symbol]{footmisc}

\makeatletter
\DeclareRobustCommand\onedot{\futurelet\@let@token\@onedot}
\def\@onedot{\ifx\@let@token.\else.\null\fi\xspace}

\def\eg{\emph{e.g}\onedot} 
\def\ie{\emph{i.e}\onedot} 
 
\def\etc{\emph{etc}\onedot}  
 
\def\etal{\emph{et al}\onedot}

\newcommand{\figcaption}[1]{\def\@captype{figure}\caption{#1}}
\newcommand{\tblcaption}[1]{\def\@captype{table}\caption{#1}}
\makeatother

\def\title#1{{\noindent\Large{\bf #1}\par}}
\def\author#1{\begin{center}{\sc #1\par}\end{center}}
\renewcommand{\section}[1]{\vspace{0.1in}\noindent{\large\bf{#1}}\par\vspace{.05in}\par\nopagebreak}

\makeatletter
\def\thebibliography#1{\section{References\@mkboth
{REFERENCES}{REFERENCES}}\list
{[\arabic{enumi}]}{\settowidth\labelwidth{[#1]}\leftmargin\labelwidth
\advance\leftmargin\labelsep
\usecounter{enumi}}
\def\newblock{\hskip .11em plus .33em minus .07em}
\sloppy\clubpenalty4000\widowpenalty4000
\sfcode`\.=1000\relax}
\makeatother

\begin{document}
\pagestyle{empty}

\title{Video Anomaly Detection for Smart Surveillance}
\author{Sijie Zhu$^{1}$, Chen Chen$^{1}$, Waqas Sultani$^{2}$ \\
${}^1$University of North Carolina at Charlotte, USA \\
${}^2$Information Technology University, Pakistan\\
\email{\{szhu3, chen.chen\}@uncc.edu}
}

\section{Related Concepts}
\begin{itemize}
\item{Anomalous event/activity detection}
\item{Novelty detection}
\end{itemize}

\section{Definition}
Anomalies in videos are broadly defined as events or activities that are unusual and signify irregular behavior.
The goal of anomaly
detection is to
temporally or spatially localize
the anomaly events in video sequences. Temporal localization (\ie indicating the start and end frames of the anomaly event in a video) is referred to as frame-level detection. Spatial localization, which is more challenging, means to identify the pixels within each anomaly frame that correspond to the anomaly event. This setting is usually referred to as pixel-level detection.

\section{Background}

In modern intelligent video surveillance systems, automatic anomaly detection through computer vision analytics plays a pivotal role which not only significantly increases monitoring efficiency but also reduces the burden on live monitoring.
Video anomaly detection has been studied for a long time, while this problem is far from being solved (as witnessed by the low accuracy on UCF-Crime  \cite{UCF} dataset) due to the difficulty of modeling anomaly events and the scarcity of anomaly data. Identifying anomaly events requires understanding of complex visual patterns, and some patterns can only be detected when long-term temporal relationship and causal reasoning are learned in the model, \eg arson, burglary, shoplifting, \etc.

Early works mainly follow the setting of general anomaly detection which may be better referred to as novelty detection, where all novel events are considered as anomaly \cite{UCSD}. This problem is typically formulated as unsupervised learning, where the models are trained with only normal video frames and validated with both normal and anomaly frames. A popular idea is to find a set of basis to represent normal frames and identify frames with high reconstruction loss or error as anomaly for inference, \eg sparse coding \cite{cong2011sparse,Avenue}, autoencoder \cite{hasan2016learning}. However, due to the limitation of data and computation, these approaches \cite{UCSD,kim2009observe,cong2011sparse,Avenue,RF} are conducted on small-scale datasets with relatively simple scenarios, which are not satisfactory for real-world surveillance applications.

While it is theoretically pleasing to consider all the novel events as anomaly, this setting has drawbacks for practical surveillance applications. Taking the campus scenario \cite{UCSD,Shanghai} as an example, riding a bike is novel (\ie considered as an anomaly) since the model only sees people walking \cite{UCSD}. However, it should not be considered as an anomaly in general for security purpose. As some anomaly activities in real world applications may have clear definitions, \eg different criminal events which follow specific patterns, recent works \cite{UCF,cleaner} start to leverage supervision for real-world anomaly detection. UCF-Crime \cite{UCF} is currently the largest anomaly detection dataset with realistic anomalies, which contains thousands of anomaly and normal videos. The training set contains both anomaly and normal videos with video-level annotation as a weak supervision, and the frame-level annotation is provided for validation set. The detection performance has been significantly improved with weakly supervised methods \cite{UCF,cleaner}. 

There is also a line of research focuses on specific anomaly detection tasks where only one type of anomaly is considered, \eg traffic accident on highway. Since the camera poses, foreground patterns, and backgrounds are highly similar and stable, the geometric prior knowledge and physics principles can be employed for manually designed detection pipelines. Several representative works \cite{CADP,traffic_1} rely on object detection to identify anomaly events.

\section{Representative Approaches}
Based on the experimental setting on the training data, video anomaly detection methods can be generally classified into three categories, \ie unsupervised, weakly-supervised, and supervised. We provide a brief overview of recent approaches for each category.

\subsection*{Unsupervised Methods}
Since real-world anomaly events happen with low probability, it is hard to capture all types of anomaly. However, normal videos are easy to access from social media and public surveillance, unsupervised methods are thus motivated to detect anomaly events with \textit{only normal videos in the training set}. Although the unsupervised methods are not able to achieve satisfactory performance on complex real-world scenarios, they are believed to have better generalization ability on unseen anomaly patterns. 

\noindent \textbf{Classic Machine Learning:} Early unsupervised methods mainly adopt classic machine learning techniques with hand-crafted features as well probability models. Kim \etal \cite{kim2009observe} propose to first extract optical flow features and find typical patterns with a mixture of probabilistic PCA (Principal Component Analysis). A space-time MRF (Markov Random Field) is then constructed to model the relationship between spatio-temporal local regions of a video for Bayesian inference. Inspired by studies of crowd behavior like social force model, Mehran \etal \cite{LDA} estimate the interaction forces in crowd to better model the normal crowd behavior. Then normal and anomaly frames are classified with BoW (Bag of Words) and LDA (Latent Dirichlet Allocation). Li \etal \cite{UCSD} introduce a mixture of DT-based (Dynamic Textures) model for temporal normalcy. And a discriminant saliency detection is utilized for measurement of spatial normalcy. Ullah \etal \cite{RF} first extract the corner features and refine them with interaction flow. A random forest is then trained to classify normal and anomaly frames. Cong \etal introduce sparse coding for anomaly detection, and Lu \etal \cite{Avenue} further propose an efficient sparse combination learning framework to achieve a speed of 150 frames per second (fps). 

\noindent \textbf{Deep Learning:}
Thanks to deep learning techniques, recent works are able to take advantage of large-scale dataset and powerful computation resource. Following the setting of unsupervised anomaly detection, a number of works \cite{hasan2016learning,luo2017remembering,luo2017revisit,gong2019memorizing} are proposed based on deep AE (autoencoder). Hasan \etal \cite{hasan2016learning} propose to learn both motion feature and discriminative regular patterns with a FCN (Fully Convolutional Network) based AE. The regularity score is computed based on the reconstruction error of AE model. To better model the temporal relationship within a video, \cite{luo2017remembering} combines FCN and LSTM (long short term memory) as a ConvLSTM-AE, which further improve the performance of AE framework. \cite{luo2017revisit} explores the combination of sparse coding and RNN (Recurrent Neural Network). A temporally-coherent sparse coding framework is proposed to introduce temporal information of video in the background of sparse coding. \cite{gong2019memorizing} proposes a memory-augmented AE to memorize prototypical normal patterns for anomaly detection. An attention-based sparse addressing is then designed to access the memory and reconstruct future frames. For all mentioned AE based methods, the anomaly events are determined based on the reconstruction error. On the other hand, \cite{ionescu2019object} proposes to formulate the problem as a multi-class classification by applying k-means clustering and one-versus-all SVM (Support Vector Machine). 

Instead of directly computing the reconstruction error of future frames with a set of basis or AE, another popular direction is to predict the future frames based on the past frames, and assign high anomaly score when the real future frame is highly different from the predicted one. To achieve future prediction, the idea of GANs (Generative Adversarial Networks) is introduced, where a generator and a discriminator are trained alternatively to achieve opposite goals. The generator is trying to produce frames that are similar to real frames, while the discriminator is trained to distinguish the generated fake frames from the real frames. With abundant training data and proper training techniques, the generator would be able to produce highly realistic fake frames which are indistinguishable from the real ones for the discriminator. Recent works \cite{Shanghai,ye2019anopcn} usually employ a FCN based framework as the generator to predict future frames. Liu \etal \cite{Shanghai} propose to add constraint on intensity, gradient and motion for future frame generation. The intensity constraint provides the consistency between generated frames and real ones on RGB space, and the gradient constraint can sharpen the generated images. The motion constraint aims to generate predicted frames with similar motions to the real ones by minimizing their optical flow difference. Ye \etal \cite{ye2019anopcn} further introduce a predictive coding module and an error refinement module based on the GAN-based framework.

\subsection*{Weakly Supervised Methods}
With the increasing video data on social media platforms such as YouTube\footnote{https://www.youtube.com/}, it is possible to access and annotate a large amount of anomaly videos \cite{UCF}. For certain application scenarios where the anomaly activities are well defined, the performance can be significantly improved by introducing supervision information. Recent works \cite{UCF,cleaner} follow the weakly supervised setting where only video-level annotation is available for training. That is the training videos are labeled with normal or anomalous; however, the temporal location of the anomaly event in each anomaly video is unknown (\ie weak supervision). 

Sultani \etal \cite{UCF} and He \etal \cite{he2018anomaly} formulate the weakly supervised problem as MIL (multiple instance learning). Every frame of a normal video should be normal, and there is at least one anomaly frame in an anomalous video. \cite{he2018anomaly} proposes a graph-based MIL framework with anchor dictionary learning, and all experiments are conducted on UCSD \cite{UCSD} dataset with a weakly supervised setting. \cite{UCF} proposes a deep learning based method along with a large-scale dataset with realistic crime-related anomalies and surveillance videos, namely UCF-Crime \cite{UCF}. A C3D framework is used to extract spatio-temporal features and generate anomaly score. To distinguish normal and anomaly frames with this weak supervision, the loss function forces the highest score of a negative video to be higher than the highest score of a normal video. With the parameters of C3D model frozen, \cite{UCF} outperforms previous works by a large margin on the UCF-Crime dataset. 

Instead of improving the MIL technique, Zhong \etal \cite{cleaner} consider the weakly supervised learning as a noisy label learning problem, where the annotation of some frames in anomaly videos are wrong. They train a GCN (Graph Convolutional Network) based cleaner to refine the noisy labels so that the classification network can be trained end-to-end with frame-level labels. 

\subsection*{Supervised Methods}
For certain scenarios where the backgrounds and objects are well defined, \eg the roads and cars for highway traffic accidents detection, recent works \cite{traffic_1,traffic_2} are usually based on the frame-level annotated training videos (\ie the temporal annotations of the anomalies in the training videos are available -- supervised setting). A popular solution is to leverage the geometric prior knowledge and object detection with additional supervision from other public datasets. 

\cite{CADP} first applies Faster-RCNN to detect vehicles, then an attention-based LSTM module is applied to learn the accident score. For recent works \cite{traffic_1,traffic_2} on AI city challenge\footnote{https://www.aicitychallenge.org/}, the frame-level annotation of accident is given on training set. Apart from applying object detection, \cite{traffic_1} models the background and space using semantic segmentation, and the geometric prior is leveraged by perspective detection. The vehicle dynamics are then represented by a spatial-temporal matrix. The anomaly events are identified based on the IOU (Intersection Over Union) of different objects while applying the NMS (Non-Maximum Suppression) procedure. \cite{traffic_2} utilizes YOLOv3 (You Only Look Once) as the object detector and specifically improves the framework for small object scenarios. Then a multi-object tracking is introduced to generate the trajectories of anomaly vehicles. The accident starting time is estimated based on a curve fitting algorithm.

\section{Datasets}
In this section, we briefly review the popular datasets for video anomaly detection. An overview of all listed datasets is provided in Table \ref{tab:1}.
\begin{table}[htbp]
    \centering
    \begin{tabular}{c|c|c|c}
    \hline
    
    \hline
    Dataset & $\#$ of Videos & Average Frames &  Example Anomalies \\
    \hline
    \hline
        UCSD Ped1 \cite{UCSD} & 70 & 201 & Bikers, small carts \\
        \hline
        UCSD Ped2 \cite{UCSD} & 28 & 163 & Bikers, small carts\\
        \hline
        Subway Entrance \cite{Subway} & 1 & 121,749  & Wrong direction, no payment\\
        \hline
        Subway Exit \cite{Subway} & 1 & 64,901 & Wrong direction, no payment\\
        \hline
        Avenue \cite{Avenue} & 37 & 839 & Run, throw, new object\\
        \hline
        UMN \cite{UMN} & 5 & 1,290 & Run\\
        \hline
        DAD \cite{DAD} & 1,730 & 100 & Traffic accidents\\
        \hline
        CADP \cite{CADP} & 1,416 & 366 & Traffic accidents\\
        \hline
        A3D \cite{yao2019unsupervised} & 1500 & 85 & Traffic accidents\\
        \hline
        DADA \cite{dada} & 2000 & 324 & Traffic accidents\\
        \hline
        DoTA \cite{dota} & 4677 & 156 & Traffic anomalies, \eg collision\\
        \hline
        Iowa DOT \cite{AI_city} & 200 & 27,000 & Traffic accidents\\
        \hline
        ShanghaiTech \cite{Shanghai}& 437 & 726 & Bikers, cars\\ 
        \hline
        UCF Crime \cite{UCF} & 1,900 & 7,247& Arson, accident, burglary, fighting\\
        \hline
        Street Scene \cite{ramachandra2020stree} & 81 & 2509 & Jaywalking, car illegally parked\\
        \hline

        \hline
    \end{tabular}
    \vspace{0.3cm}
    \caption{An overview of datasets for video anomaly detection.}
    \vspace{-0.3cm}
    \label{tab:1}
\end{table}


\textbf{UCSD:} 
The UCSD dataset contains two subsets, denoted as Ped1 and Ped2. They are captured with different camera poses at two spots in UCSD campus where most pedestrians walk. The training set (34 clips for Ped1 and 16 clips for Ped2) only contains normal frames, and the test set (36 clips for Ped1 and 12 clips for Ped2) consists of both normal and anomaly frames. Frame-level annotation is provided for all test clips and 10 of them have pixel-level ground-truth. UCSD dataset considers pedestrians walking as the normal pattern, so non pedestrian entities like bikers and skaters are defined as anomaly instances. Dataset link: \url{http://www.svcl.ucsd.edu/projects/anomaly/dataset.html}

\textbf{Subway:}
Subway \cite{Subway} dataset contains two subsets, \ie Subway Entrance and Subway Exit. They contain only one long surveillance video each in subway station. They are first proposed specifically for real-time detection of unusual events detection in crowded subway scenes, \eg moving in the wrong direction, or no payment. Dataset link: \url{http://vision.eecs.yorku.ca/research/anomalous-behaviour-data/}

\textbf{Avenue:}
The Avenue \cite{Avenue} dataset contains 15 videos, and each video is about 2 minutes long. The total frame number is 35,240. 8,478 frames from 4 videos are used as training set. Typical unusual events include running and throwing objects. Dataset link: \url{http://www.cse.cuhk.edu.hk/leojia/projects/detectabnormal/dataset.html}

\textbf{UMN:}
The UMN \cite{UMN} (University of Minnesota) dataset consists of five videos captured from different angles. The normal pattern is defined as walking and the main anomaly activity is running. Dataset link: \url{http://mha.cs.umn.edu/}

\textbf{DAD:}
DAD \cite{DAD} (Dashcam Accident Dataset) is proposed specifically for accident detection. The normal pattern is vehicles moving around and anomaly events include different traffic accidents, \eg car hits car, or motorbike hits motorbike. DAD dataset consists of 678 videos from six cities. 58 videos are used for training. For the rest 620 videos, 620 clips with accidents are sampled as positive clips and 1130 normal clips are sampled as negative clips. They are then randomly split into two subsets, \ie 455 positive and 829 negative clips for training, and 165 positive and 301 negative clips for testing. Dataset link: \url{https://aliensunmin.github.io/project/dashcam/}

\textbf{CADP:}
CADP \cite{CADP} (Car Accident Detection and Prediction) focuses on car accident on CCTV (Closed-Circuit Television) cameras. All the 1416 videos of CADP contain traffic accidents, and 205 of them have temporal as well as spatial annotations. CADP contains videos captured under various camera types, qualities, weather conditions, and the anomaly events are realistic for real-world applications. Dataset link: \url{https://ankitshah009.github.io/accident_forecasting_traffic_camera}

\textbf{A3D:}
A3D \cite{yao2019unsupervised} consists of 1500 on-road abnormal event video clips from dashboard cameras. Each video contains an
abnormal traffic event, and the anomaly start and end times are annotated by human annotators. A total of 128,175
frames (ranging from 23 to 208 frames) at 10 frames per second are clustered into 18 types of traffic accidents.
Dataset link: \url{https://github.com/MoonBlvd/tad-IROS2019}

\textbf{DADA:}
DADA \cite{dada} is a traffic accident dataset collected for driver attention prediction in accidental scenarios. It has 658,476 available frames contained in 2000 videos with the resolution of 1584$\times$660. The videos are divided into 54 kinds of categories, such as ``hitting" and ``out of control", based on the participants of accidents
(\eg pedestrian, vehicle, cyclist, etc.). The
spatial crash-objects, temporal window of the occurrence of accidents are annotated. 
Dataset link: \url{https://github.com/JWFangit/LOTVS-DADA}

\textbf{DoTA:}
DoTA \cite{dota} (Detection of Traffic Anomaly) is a recent traffic anomaly detection dataset containing 4,677 videos with temporal, spatial, and categorical annotations. The objective is to introduce a when-where-what pipeline to detect, localize, and recognize anomalous events from egocentric videos.
The video clips are collected from YouTube channels with diverse dash camera accident videos from different countries under different weather and lighting conditions. Dataset link: \url{https://github.com/MoonBlvd/Detection-of-Traffic-Anomaly}

\textbf{Iowa DOT Traffic:}
Iowa DOT (Department of Transportation) Traffic dataset \cite{AI_city} consists of 200 videos, each approximately 15 minutes in length, recorded at 30 fps and $800 \times 410$ resolution. Training and testing set each contains 100 videos. As the official dataset for the 2018 AI City challenge \cite{AI_city} Track 3, it does not provide annotation for the testing set. Main anomaly patterns are car crashes and stalled vehicles. Dataset link: \url{https://www.aicitychallenge.org/2018-ai-city-challenge/}

\textbf{ShanghaiTech:}
ShanghaiTech \cite{Shanghai} dataset is collected in ShanghaiTech University under 13 scenes with complex light conditions and camera viewpoints. It consists of 437 videos with 726 average frames each. The training set 
consists of 330 normal videos and testing set contains 107
videos with 130 anomalies. Anomaly events include unusual patterns in campus such as bikers or cars. Dataset link: \url{https://svip-lab.github.io/dataset/campus_dataset.html}

\textbf{UCF Crime:}
UCF Crime \cite{UCF} consists of 1900 untrimmed videos covering 13 real-world anomaly events, including Abuse, Arrest, Arson, Assault, Road Accident, Burglary, Explosion, Fighting, Robbery, Shooting, Stealing, Shoplifting, and Vandalism. 950 of them are normal videos and the rest videos contain at least one anomaly event for each. The training set contains 800 normal and 810 anomalous videos. The remaining 150 normal and 140 anomalous videos are  temporally annotated for validation. Both training and testing sets cover all the 13 anomaly events. Some of the videos may contain multiple anomaly categories, \eg robbery along with fighting, burglary with vandalism, arrest with shooting. All the videos are realistic for real-world surveillance applications. Furthermore, UCF Crime covers different light conditions, image resolutions, and camera poses in complex scenarios, thus is very challenging. Dataset link: \url{https://www.crcv.ucf.edu/projects/real-world/}

\textbf{Street Scene:}
Street Scene \cite{ramachandra2020stree} dataset is focused on single scene anomaly detection. It consists of 46 training videos and 35 testing videos taken from a static USB camera looking down on a scene of a two-lane street with bike lanes and pedestrian sidewalks.
There are a total of 203,257 color video frames (56,847 for training and 146,410 for testing) with 1280 $\times$ 720 resolution. The frames were extracted from the original videos
at 15 frames per second. 17 types of anomaly events/activities are presented in the dataset such as jaywalking, loitering, car illegally parked, etc. Dataset link: \url{https://www.merl.com/demos/video-anomaly-detection}

\section{Benchmarks}
In this section, we introduce popular evaluation metrics and show existing results on five popular benchmark datasets, \ie UCSD Ped2 \cite{UCSD}, Avenue \cite{Avenue}, UMN \cite{UMN}, Shanghai Tech \cite{Shanghai}, UCF \cite{UCF}, and Iowa DOT Traffic \cite{AI_city}. 

The frame-level evaluation criterion uses the frame-level ground truth annotations to determine which
detected frames are true positives (\ie true anomaly frames) and which are false positives, yielding frame-level true positive and false positive rates. In pixel-level evaluation, it requires the algorithm to take into account the spatial locations of anomaly objects in frames. A detection is considered to be correct if it covers at least 40\% of anomaly pixels in the ground-truth \cite{UCSD}. The pixel-level evaluation can be conducted only if the pixel-level annotations are available for the testing videos. 

As shown in Tables \ref{tab:table_2} and \ref{tab:table_3}, the frame-level AUC (Area Under the Curve) of ROC (Receiver Operating Characteristic) curve is widely used as the evaluation metric for temporal localization of anomaly events. Since the anomaly detection can be considered as a binary classification for each frame, the ROC curve is generated by applying different thresholds for the anomaly score of each frame and calculating the TPR (True Positive Rate) and FPR (False Positive Rate). 
\begin{table}[htbp]
    \centering
    \begin{tabular}{c|c|c|c|c}
    \hline
    
    \hline
       Method  & UMN & UCSD Ped2 & Avenue & Shanghai Tech \\
       \hline\hline
       Mehran \etal \cite{LDA} &96.0 & -& -& -\\
       \hline
       Cong \etal \cite{cong2011sparse} & 97.8 & - & - & -\\
       \hline
       Li \cite{UCSD,luo2017revisit}& 99.5 & 69.3& -&-\\
       \hline
       Hasan \etal \cite{hasan2016learning} & - & 90.0 & 70.2& -\\
         \hline
        Luo \etal \cite{luo2017revisit} &-& 92.21 & 81.71 & 68.0\\
         \hline
        Gong \etal \cite{gong2019memorizing} & - & 94.1 & 83.3 & 71.2\\
        \hline
         Liu \etal \cite{Shanghai}& - & 95.4 & 85.1 & 72.8\\
         \hline 
          Ye \etal \cite{ye2019anopcn} & - & 96.8 & 86.2 & 73.6\\
         \hline
         Ionescu \etal \cite{ionescu2019object} & 99.6 & 97.8 & 90.4 & 84.9\\
         \hline
         
         \hline
    \end{tabular}
    \vspace{0.3cm}
    \caption{Frame-level anomaly detection evaluation. AUC (\%) of existing works on UCSD, UMN, Avenue, and Shanghai Tech with unsupervised setting.}
    \vspace{-0.3cm}
    \label{tab:table_2}
\end{table}

\begin{table}[htbp]
    \centering
    \begin{tabular}{c|c|c|c}
    \hline
    
    \hline
    Method     &  UCSD & Shanghai Tech & UCF Crime\\
    \hline\hline
    He \etal \cite{he2018anomaly} & 90.1 & -& - \\ 
    \hline
    Sultani \etal \cite{UCF} & -& -& 75.41\\
    \hline
    Zhong \etal \cite{cleaner} & 93.2& 84.44 & 82.12\\
    \hline
    
    \hline
    \end{tabular}
    \vspace{0.3cm}
    \caption{AUC (\%) of existing works on UCSD, Shanghai Tech, UCF Crime with weakly supervised setting.}
    \vspace{-0.3cm}
    \label{tab:table_3}
\end{table}

For traffic accident detection with supervised setting, the F1 score, RMSE (Root Mean Square Error) of anomaly start time, and S3 score are used as evaluation metrics. The F1 score is defined as:
\begin{equation}
    F1 = \frac{2TP}{2TP+FP+FN},
\end{equation}
where TP, FP, and FN denote true positive, false positive, and false negative numbers. The S3 score is computed as:
\begin{equation}
    S3=F1(1-NRMSE),
\end{equation}
where the NRMSE denotes the normalized root mean square error \cite{AI_city}. We show the performance of two state-of-the-art methods on Iowa DOT Traffic dataset in Table \ref{tab:table_4}.

\begin{table}[htbp]
    \centering
    \begin{tabular}{c|c|c|c}
    \hline
    
    \hline
    Method & F1 score & RMSE & S3 score \\
    \hline\hline
     UWIPL \cite{traffic_2} &0.9577 & 6.7461 & 0.9362 \\
     \hline
     Traffic Brain \cite{traffic_1}  & 0.9706 & 5.3058 & 0.9534 \\
      \hline
      
      \hline
    \end{tabular}
    \vspace{0.3cm}
    \caption{F1 score, RMSE, and S3 score of two top-performing methods on Iowa DOT Traffic dataset with supervised setting.}
    \vspace{-0.3cm}
    \label{tab:table_4}
\end{table}

\section{Video Anomaly Detection Conference Workshop}
The NVIDIA AI City Challenge\footnote{https://www.aicitychallenge.org/} was launched in 2017 and has been held as a full-day workshop of IEEE/CVF Conference on Computer Vision and Pattern Recognition (CVPR) since 2018. Traffic anomaly detection in surveillance videos is one track of the challenge.

\section{Open Problems}
Although there has been significant progress towards building efficient video anomaly detection algorithms in recent years, in particular the deep learning-based approaches, we highlight a few possible open problems that are worth exploring in the future.
\begin{itemize}
    \item Although different learning frameworks have been adopted, the learned representations are still not satisfactory for distinguishing complex anomaly activities. Possible better representations include better 3D feature extractor, attention mechanism, and causal reasoning (identifying the cause of an anomaly event, \eg too fast $\xrightarrow{}$ accidents).
    \item Early works mainly focus on the unsupervised setting, and recent works have shown potential on improving performance by leveraging some supervision information for certain scenarios. It would be promising to explore better setting for practical applications, \eg better trade-off between the generalization ability (unsupervised setting) and performance (weakly supervised setting).
    \item It may be acceptable for anomaly detection systems operating in public spaces where there is no expectation of privacy. However, what if the technology needs to be applied to non-public spaces where there is a stronger expectation of privacy? It is worth exploring effective ways to de-identify the training videos and train anomaly models with de-identified data.

    \item
    The current anomaly detection approaches or systems act as an alerting mechanism. How do we explain the AI decisions and convey these effectively to stakeholders, \eg law enforcement, attorneys, media, local residents, and broader community. We expect techniques to close the gap between performance and interpretable AI models.


\end{itemize}

\nocite{*}
\bibliographystyle{plain}
\bibliography{template}
\end{document}